# Dataset Properties Shape the Success of Neuroimaging-Based Patient Stratification: A Benchmarking Analysis Across Clustering Algorithms

**Authors:** Yuetong Yu[1,2], Ruiyang Ge[1,2], Ilker Hacihaliloglu[1,3,4], Alexander Rauscher[1,5], Roger Tam[1,6], Sophia Frangou[1,2,7]

1.Djavad Mowafaghian Centre for Brain Health, University of British Columbia, Canada
2. Department of Psychiatry, University of British Columbia, Canada
3. Department of Radiology, University of British Columbia, Canada
4. Department of Medicine, University of British Columbia, Canada
5. Department of Pediatrics, University of British Columbia, Canada
6. School of Biomedical Engineering, University of British Columbia
7. Department of Psychiatry, Icahn School of Medicine at Mount Sinai, USA

**Abstract**

Background: Data driven stratification of patients into biologically informed subtypes holds promise for precision neuropsychiatry, yet neuroimaging-based clustering methods often fail to generalize across cohorts. While algorithmic innovations have focused on model complexity, the role of underlying dataset characteristics remains underexplored. We hypothesized that cluster separation, size imbalance, noise, and the direction and magnitude of disease-related effects in the input data critically determine both within-algorithm accuracy and reproducibility.

Methods: We evaluated four widely used stratification algorithms—HYDRA, SuStaIn, SmileGAN, and SurrealGAN—on a suite of synthetic brain-morphometry cohorts derived from the Human Connectome Project–Young Adult dataset. Three global transformation patterns (monotonic increase, monotonic decrease, mixed change) were applied to 600 pseudo-patients against 508 controls, followed by four within-dataset variations (Widespread, Localized, Increased Noise, Subtle) varying cluster count (k = 2–6), overlap, and effect magnitude. Algorithm performance was quantified by accuracy in recovering the known ground-truth clusters.

Results: Across 122 synthetic scenarios, data complexity consistently outweighed algorithm choice in predicting stratification success. Well-separated clusters yielded high accuracy (> 90 %) for all methods, whereas overlapping, unequal-sized, or subtle effects reduced accuracy by up to 50 %. SuStaIn could not scale beyond 17 features, HYDRA's accuracy (45 %–99 %) varied unpredictably with data heterogeneity. SmileGAN and SurrealGAN maintained robust pattern detection but did not assign discrete cluster labels to individuals.

Conclusions: The study results demonstrate the impact of statistical properties of input data across algorithms and highlight the need for using realistic dataset distributions when new algorithms are being developed and suggest great focus on data - centric strategies that actively shape and standardize the input distributions.

## INTRODUCTION

The diversity of manifestations of human illness — manifesting as diverse symptom profiles, courses, and treatment responses — poses a fundamental obstacle to defining biologically meaningful patient subgroups or “biotypes” that translate to clinical practice (Feczko & Fair, 2020; Nunes et al., 2020). This challenge is particularly acute in neuropsychiatric disorders, where symptom-defined categories often fail to map onto underlying pathophysiology (Hyman, 2010). Nevertheless, the proliferation of large‑scale, multidomain repositories together with continual advances in computational methods has invigorated efforts to uncover reproducible, data‑driven subtypes (Gao et al., 2023; Feczko & Fair, 2020). These advancements have been particularly impactful in the data-driven analysis of neuroimaging data, where automated machine learning algorithms have become leading tools for identifying patient biotypes.

Automated machine‑learning techniques, ranging from support-vector machines to deep generative models, now underpin much of contemporary neuroimaging stratification in brain disorders, allowing investigators to detect subtle, high-dimensional patterns in structure and connectivity with minimal a priori assumptions (Ezugwu et al., 2021; Singh et al., 2024). Despite this methodological progress, however, few proposed biotypes have achieved consensus or clinical uptake. Poor reproducibility is the primary challenge as clusters identified in one cohort often fail to replicate in others, and the specific neuroimaging features driving stratification can differ markedly across studies (Beijers et al., 2020; Wen et al., 2024). Such inconsistencies are typically ascribed to differences in algorithmic assumptions and implementations (Gao et al., 2023; Singh et al., 2024) which have spurred the development of even more sophisticated approaches aimed at enhancing performance.

We contend that a more fundamental impediment may lie in the statistical properties of the input data themselves, which vary substantially between real-world cohorts but are seldom accounted for in algorithm development. This issue is further compounded by the reliance on synthetic datasets for the development of new machine learning methods. Most new methods are initially validated on synthetic datasets that assume tidy, multivariate normality or clear-cut cluster separation, whereas observed neuroimaging data often exhibit partial overlap, skewness, and site- or scanner-related batch effects (Dalmaijer et al., 2022). This disconnect can engender an optimistic view of algorithm performance and underlie the poor out-of-sample generalization that plagues current approaches.

Rather than making algorithm choice the centerpiece of investigation, this study foregrounds the impact of dataset characteristics on stratification outcomes. To illustrate this point, we used four widely used approaches in neuroimaging: HeterogeneitY through DiscRiminant Analysis (HYDRA) (Valor et al., 2017), Subtype and Stage Inference (SuStaIn) (Young et al., 2018), SeMI-supervised cLustEring-Generative Adversarial Network (SmileGAN) (Yang et al., 2021), and Semi-SUpeRvised ReprEsentAtion Learning via GAN (SurrealGAN) (Yang, 2022). In terms of data types, we focus on brain morphometric measures (cortical thickness, surface area, subcortical volumes) chosen for their widespread availability and proven sensitivity to developmental and degenerative changes (Fjell & Walhovd, 2010; Raz et al., 2010). However, a central strength of our approach lies in its modality-agnostic architecture: the same analytical pipeline can be readily applied to measures of diffusion or connectivity, and even to non-imaging data domain such as behavioural, clinical, multi-omic or electrophysiological measures. Accordingly, insights from this study into how dataset characteristics influence stratification performance are broadly generalizable across diverse data domain.

To test the influence of dataset characteristics on algorithmic performance, we constructed a series of 122 of synthetic cohorts that systematically vary cluster count and size, and the degree of overlap between patient and control distributions. This controlled evaluation permits us to quantify within‐algorithm sensitivity to data heterogeneity as well as between-algorithm differences in resilience to non-ideal conditions. By centering on input properties, this study aims to provide clear, empirical guidance for selecting and responsibly applying stratification tools in neuroimaging research and beyond.

## METHODS

### Algorithms

All algorithms have been previously published and are described briefly below and in greater detail in the Supplementary Information.

**Heterogeneity through Discriminative Analysis** (HYDRA) is a semi-supervised method that combines binary classification and unsupervised clustering to uncover subpopulations within patient cohorts (Varol et al., 2017). After mapping neuroimaging features into spatially defined “parcels” (e.g., atlas-derived regions or data-driven components), HYDRA constructs a convex polytope—an ensemble of maximum-margin hyperplanes—that encloses the patient data while separating it from healthy controls. Determinantal point processes ensure diverse initialization for multiple runs, and a consensus step aggregates these into stable subject-level clusters.

**Semi-supervised Clustering via Generative Adversarial Networks (SmileGAN)** employs a GAN framework to model patient deviations from control distributions (Yang et al., 2021). Patient data are first z-scored against healthy references, then a generator learns mappings that produce realistic pseudo-patient samples, while a discriminator enforces fidelity via adversarial loss. Regularization terms—sparsity, Lipschitz continuity, and explicit clustering losses—enhance interpretability and reduce noise, and an inverse mapping assigns real patients to the discovered clusters.

**Semi-Supervised Representation Learning via GAN (SurrealGAN)** decomposes disease-related effects into low-dimensional R-indices that capture independent deviation patterns (Yang & Roy, 2022). A forward transformation synthesizes pseudo-patient data from controls, and an inverse network extracts R-indices while enforcing monotonicity, orthogonality, continuity, and sparsity. These indices can subsequently be used to define homogeneous patient subgroups, even though the model itself does not assign discrete cluster labels.

**Subtype and Stage Inference (SuStaIn)** infers both categorical subtypes and their temporal progression from cross-sectional data by modeling disease as an ordered sequence of z-score “events” (Young et al., 2018). Patient measures are z-normalized to controls, and an event-based generative model assumes a monotonic ordering of threshold crossings across parcels. Markov Chain Monte Carlo sampling estimates patient probabilities for each subtype and stage, enabling simultaneous clustering and staging of disease progression.

## Datasets

The use of multiple synthetic datasets offers several advantages. First, it creates a fully controlled environment in which parameters—such as cluster separation or effect-size distributions—can be precisely varied to isolate their impact on algorithm performance (Lu et al., 2023). Second, it allows systematic testing of robustness across hypothetical scenarios—for example, simulating varying patterns of cortical thinning to evaluate algorithm sensitivity to subtle versus pronounced neuroanatomical changes. Finally, because synthetic data evade privacy and access restrictions, they support unrestricted sharing, complete transparency, and exact replication across research groups.

### *Synthetic Datasets From Prior Studies*

We selected two pre-existing synthetic datasets, which we termed Syn1 and Syn2 because they were utilized in the development of SmileGAN (Yang et al., 2021) and SurrealGAN (Yang et al., 2022). This choice enables direct reference to the findings reported in the corresponding studies. Syn1 comprised 1,200 participants, each contributing 145 variables, of whom 600 were designated as healthy controls and 600 as pseudo-patients. Syn2 included 1,500 participants, each contributing 100 variables, of whom 600 were designated as the control group and 900 the pseudo-patient group. In each dataset, the values of the contributing variables were generated by random sampling from a normal distribution with a mean of 1 and a standard deviation of 0.1. The control group in both datasets was designed to be used as a reference population. Pseudo-patient data in both datasets were separated into three equal-sized clusters, each representing a distinct disease pattern of random reductions in variable values. Further details on Syn1 and Syn2 are provided in the Supplementary Information.

### *Generation of Controlled Synthetic Cohorts*

To systematically explore how dataset properties affect stratification algorithms, we began with the real-world morphometric data of 1,108 healthy participants from the Human Connectome Project-Young Adult (HCP-YA) cohort (https://www.humanconnectome.org/study/hcp-young-adult). All $T_1$-weighted images were processed with FreeSurfer image analysis suite (http://surfer.nmr.mgh.harvard.edu/) to extract 150 regional measures (subcortical volumes via the Aseg atlas and cortical thickness & surface area via the Desikan–Killiany atlas). The workflow (Figure 1) for generating the controlled synthetic datasets comprised the following steps:

***(1) Defining a healthy reference and a pseudo-patient subgroup***: From the HCP-YA cohort, 508 participants were randomly assigned to serve as a healthy reference group. The values of the brain morphometric measures in this group were unaltered throughout all analyses. The remaining 600 were designated as pseudo-patients whose data were systematically manipulated in subsequent steps to simulate disease-related variation.

**(2) *General Framework for Pseudo-patient data transformation***: The framework applied ensured that the synthetic manipulations of the values of each individual's 150 variables (i.e., regional morphometric measure) were proportional to the original variable values and tuned by $\sigma$ and $\alpha$. Specifically, for each pseudo-patient $i$ and each variable $j$, the original value $v_{ij}$ was transformed to a synthetic value $tv_{ij}$ according to the following transformation formula:

$$tv_{ij} = v_{ij} \pm v_{ij} * s_{ik} * N(1, \sigma) * \alpha$$

where: $s_{ik}$ = 3-dimensional severity vector sampled from a multivariate uniform distribution $[0,1]^3$ (distinct for each cluster $k$); $N(1, \sigma)$ = Gaussian noise with mean 1 and SD $\sigma$ and $\alpha$ = scaling factor controlling the magnitude of change.

**(3) *Three Global Transformation Patterns:*** We generated three core datasets, Syn3, Syn4 and Syn5, each differing in the direction of morphometric change:

- Syn3 simulated monotonic increase: $tv_{ij} = v_{ij} + v_{ij} * s_{ik} * N(1, \sigma) * \alpha$
- Syn4 simulated monotonic decrease: $tv_{ij} = v_{ij} - v_{ij} * s_{ik} * N(1, \sigma) * \alpha$
- Syn5 simulated mixed random increases and decreases: $tv_{ij} = v_{ij} \pm v_{ij} * s_{ik} * N(1, \sigma) * \alpha$

**(4) *Four Core Within-Dataset Variations***

Each of the three synthetic datasets, Syn3, Syn4, and Syn5, served as the basis for additional within-dataset variations designed to simulate different patterns of morphometric alteration among pseudo-patients. There are four main variations that were instantiated across five cluster-count scenarios, with k set to 2, 3, 4, 5, or 6 pseudo-patient clusters. The size of the clusters was also varied to include both equal and unequal clusters sizes, allowing for a comprehensive evaluation of how both variation type and cluster complexity influence algorithmic performance (Table S1).

- **Widespread:** This variation applied broad distortions across a relatively large number of regions per cluster, for a cluster range of 2-6. Specifically, 21 variables were altered per cluster, with six variables overlapping across clusters. This variation used a noise level of $\sigma$ = 0.05 and a scaling factor of $\alpha$ = 0.30.
- **Increased Noise**: This variation followed the same structure as the widespread condition (21 variables per cluster; 2-6 clusters) but increased the level of measurement noise to $\sigma$ = 0.20. In this context, "noise" refers to the introduction of random variability to the morphometric data of the pseudo-patient groups to simulate real-world variability in neuroimaging data, where measurements are subject to natural biological differences, scanner artifacts, preprocessing inconsistencies, and other uncontrolled sources of variation.

- **Subtle**: This variation also followed the widespread structure (21 variables per cluster; 2-6 clusters) but reduced the scaling factor to $\alpha$ = 0.20 while maintaining a low noise level ($\sigma$ = 0.05), thereby simulating more nuanced, hard-to-detect changes in brain structure.
- **Localized:** This variation focused on focal perturbations, altering only six variables per cluster with no overlap between clusters. The noise and scaling parameters remained the same as in the widespread condition ($\sigma$ = 0.05, $\alpha$ = 0.30), enabling a controlled comparison of extent versus intensity.

The synthetic datasets generated for this study can be accessed at https://github.com/YuYT98/clustering_paper_data.

### Algorithm performance

The performance of each algorithm on each synthetic dataset was evaluated by comparing its results to the ground truth, which refers to the predefined patterns intentionally embedded in the synthetic datasets during their creation. Accuracy was used as the primary metric to assess how well each algorithm identified these predefined patterns, providing a clear measure of the algorithm's ability to recover the known structure of the data.

## RESULTS

### SuStaIn

In the current study where the sample size of the synthetic datasets was over 1,000, the SuStaIn algorithm was able to process a maximum of 17 variables, far below the over 100 variables included in each synthetic dataset considered. As a result, the algorithm was unable to perform in any of the synthetic datasets analyzed. SuStaIn is fundamentally based on ordering variables into a disease progression model. When too many input variables are included, the search space of possible progression sequences grows exponentially. With 17+ variables, the number of possible permutations of disease progression patterns becomes extremely large, making it difficult for the Expectation-Maximization (EM) algorithm to efficiently explore the solution space. SuStaIn uses a Bayesian inference framework combined with EM to estimate disease progression patterns. The EM algorithm relies on iterating between the Expectation step (Estimating latent disease stages for each subject and the maximization step (Updating model parameters based on estimated stages). As the number of variables increases, the maximization step becomes increasingly complex because it must optimize across a much larger number of progression sequences, leading to poor convergence or local minima. Additionally, higher number of variables leads to high parameter variance, making the algorithm output sensitive to initial conditions.

**HYDRA**

The accuracy of HYDRA for Syn1 and Syn2 was 0.99 and 0.52 respectively. Its accuracy in all other datasets varied from 0.45-0.99 (Figure 2). Notably, the performance of HYDRA across the different synthetic databases was highly variable in a manner that could not be accounted for by the characteristics of the data distribution, presenting a perplexing challenge.

**SmileGAN**

SmileGAN achieved an accuracy of 0.99 in Syn1 and 0.61 in Syn2. Its accuracy in all other datasets varied from 0.37-0.99 (Figure 3). It achieved its highest accuracy in Syn1 and maintained high accuracy in the Syn3 dataset variations with 2 pseudo-patient clusters of equal size. Its performance dropped for variations with greater complexity in terms of cluster numbers and size, and degree or direction of change.

**SurrealGAN**

SurrealGAN demonstrated an accuracy of 0.97 in Syn1 and 0.83 in Syn2. Its accuracy in all other datasets varied from 0.38-0.99 (Figure 4) if available. It achieved its highest accuracy 0.99 in Syn3 and maintained reasonable accuracy in across all variations of the synthetic datasets with a notable drop in performance in the most complicated dataset Syn5, particularly with higher cluster numbers.

**Pattern identification versus individual-level classification**

The key difference between HYDRA and the GAN-based methods (SmileGAN and SurrealGAN) lies in the nature of their outcomes: HYDRA identifies clusters at the level of individual patients, whereas SmileGAN and SurrealGAN identify patterns of disease profiles which can be shared by individual patients. SmileGAN and SurrealGAN operate similarly, but they use different techniques and regularization levels to improve subtype detection.

It was important to assess whether the disease profiles identified by these two algorithms were sufficiently distinct at the individual-patient level, as this distinction is critical for the translational value of these outputs. Clinical decision-making ultimately occurs at the level of an individual patient, making it essential to determine whether the algorithm-generated disease patterns could meaningfully stratify the pseudo-patients in the synthetic datasets. We focus on SurrealGAN outputs given its superior performance in previous analyses. In SurrealGAN, when the algorithm identifies k R-indices, each patient is represented as a vector of $k$ variables, where each variable quantifies the degree to which the patient's profile aligns with each R-index. We applied K-means clustering to these vectors from all the synthetic datasets (Syn3-5 and their variations) to assess whether a conventional, centroid-based method would yield well-separated disease clusters based on the R indices identified by SurrealGAN. Figure 5 showcases the results for Syn4 as an example as the pattern was similar for all synthetic datasets.

**Computational Environment and Efficiency**

The computational environment for all reported clustering results utilized the same capacity (8 CPUs, 1 GPU, 16G Memory) of UBC ARC Sockeye, a centralized supercomputing infrastructure [https://arc.ubc.ca/compute-storage/ubc-arc-sockeye]. SuStaIn failed to converge with more than 17 variables even when we increased computational resources to 64 CPUs and 186GB of RAM memory, even after running for over 168 hours, indicating significant limitations in computational efficiency. SurrealGAN and SmileGAN provided results between 2 to 22 hours, depending on the number of clusters and the complexity of the dataset. HYDRA completed clustering tasks within much shorter time frames, usually within 10 minutes, but as previously noted it was not able to resolve the underlying clusters in some of datasets.

## DISCUSSION

This study systematically evaluated how dataset properties and algorithmic design jointly influence the identification of disease related patterns from neuroimaging features. By leveraging multiple synthetic datasets, we show that data complexity—cluster separation, size imbalance, noise, magnitude and direction of change—exerts a stronger influence on both within-algorithm accuracy and between-algorithm reproducibility than the choice of method itself.

SuStaIn was unable to process datasets with more than 17 features at moderate sample sizes, reflecting its limited scalability and dependence on a small, investigator-defined feature set. While selecting variables based on prior knowledge can ensure inclusion of well-established markers, it risks overlooking subtler or novel patterns of heterogeneity. Moreover, SuStaIn's reliance on pre-specified z-scores—rather than data-driven thresholds—undermines its validity and generalizability (Hawkins et al., 2018).

HYDRA demonstrated the capacity to stratify individuals in nearly all scenarios, yet its performance was unpredictably variable. Its convex-polytope classification appears vulnerable to high data heterogeneity (Kulesza & Taskar, 2012), raising concerns about reproducibility in real-world cohorts with unknown distributions. Nevertheless, HYDRA's ability to assign patient-level clusters remains a critical advantage for potential clinical application, provided that strategies are developed to stabilize its outputs across variable data conditions.

SmileGAN and SurrealGAN, as generative-adversarial-network approaches, excelled at uncovering shared disease patterns. SurrealGAN in particular maintained robustness under unequal cluster sizes and elevated noise, reflecting its enhanced regularization. However, neither GAN produced hard cluster memberships; instead, individuals were mapped onto continuous "disease spectra," complementing—but not replacing—categorical clustering methods like HYDRA.

Our synthetic benchmarks underscore that well-separated clusters yield high accuracy across all methods, whereas overlapping, unequal-sized, or subtle alterations markedly degrade performance. Prior synthetic studies often assume idealized, Gaussian-like clusters (Dalmaijer et al., 2022), failing to capture real-world complexities such as measurement noise or small but clinically meaningful subgroups. We show that unequal cluster sizes introduce a systematic bias, as smaller clusters are frequently absorbed into larger ones (Finch, 2019), and that increased noise disproportionately encumber algorithms lacking robust regularization. Subtle effect sizes demand large samples and sensitive methods, since small changes may fall below detection thresholds. These insights highlight that algorithm selection must be guided by an understanding of dataset properties, and that addressing size imbalance and noise through large-scale data collection (Atif et al., 2024), advanced preprocessing, and algorithmic adaptations (e.g., density-aware clustering) is essential for reliable subtype discovery.

The implications for patient stratification are threefold. First, algorithms should undergo rigorous pre-deployment testing on synthetic cohorts that reflect expected complexities—cluster overlap, size imbalance, noise, and subtlety—to reveal potential failure modes before clinical application (Lu et al., 2023). Second, methodological transparency is necessary—through public sharing of synthetic generation scripts and performance metrics—for exact replication and fair comparisons across research groups. Third, the field may benefit from exploring more data-centric methods, for example, strategies aimed at creating realistic variations of existing cases via generative models—to ensure that algorithms see the full range of patterns they might encounter in practice. Another option is domain adaptation, where models are trained to recognize and adjust to differences in data distributions, making them more robust when applied to new cohorts. We can also integrate meta-learning or ensemble frameworks that automatically tune or combine multiple algorithms based on simple data-quality indicators (like variance, overlap, or imbalance).

While our focus was morphometry, these conclusions generalize to other modalities (diffusion, connectivity, genomics, electrophysiology) facing similar data-complexity challenges.

Limitations of this work include the inherent gulf between synthetic and truly in vivo data; future efforts should explore hybrid approaches that blend synthetic perturbations with real patient datasets. Expanding the algorithmic repertoire to include density-based, graph-theoretic, or model-based clustering could yield further insights. Finally, we evaluated performance primarily by accuracy; subsequent studies should incorporate metrics of stability, interpretability, and clinical utility (e.g., prognostic validity).

**Conclusion**

Stratifying patients into biologically meaningful subtypes is essential for advancing personalized psychiatry. Our systematic evaluation reveals that dataset properties often overshadow algorithmic nuances in driving performance and may require further focus on datacentric strategies. These principles extend well beyond neuroimaging, offering a blueprint for robust subtype discovery across medical research.

**ACKNOWLEDGMENTS**

This research was supported in part through computational resources and services provided by Advanced Research Computing at the University of British Columbia.

**FIGURE**

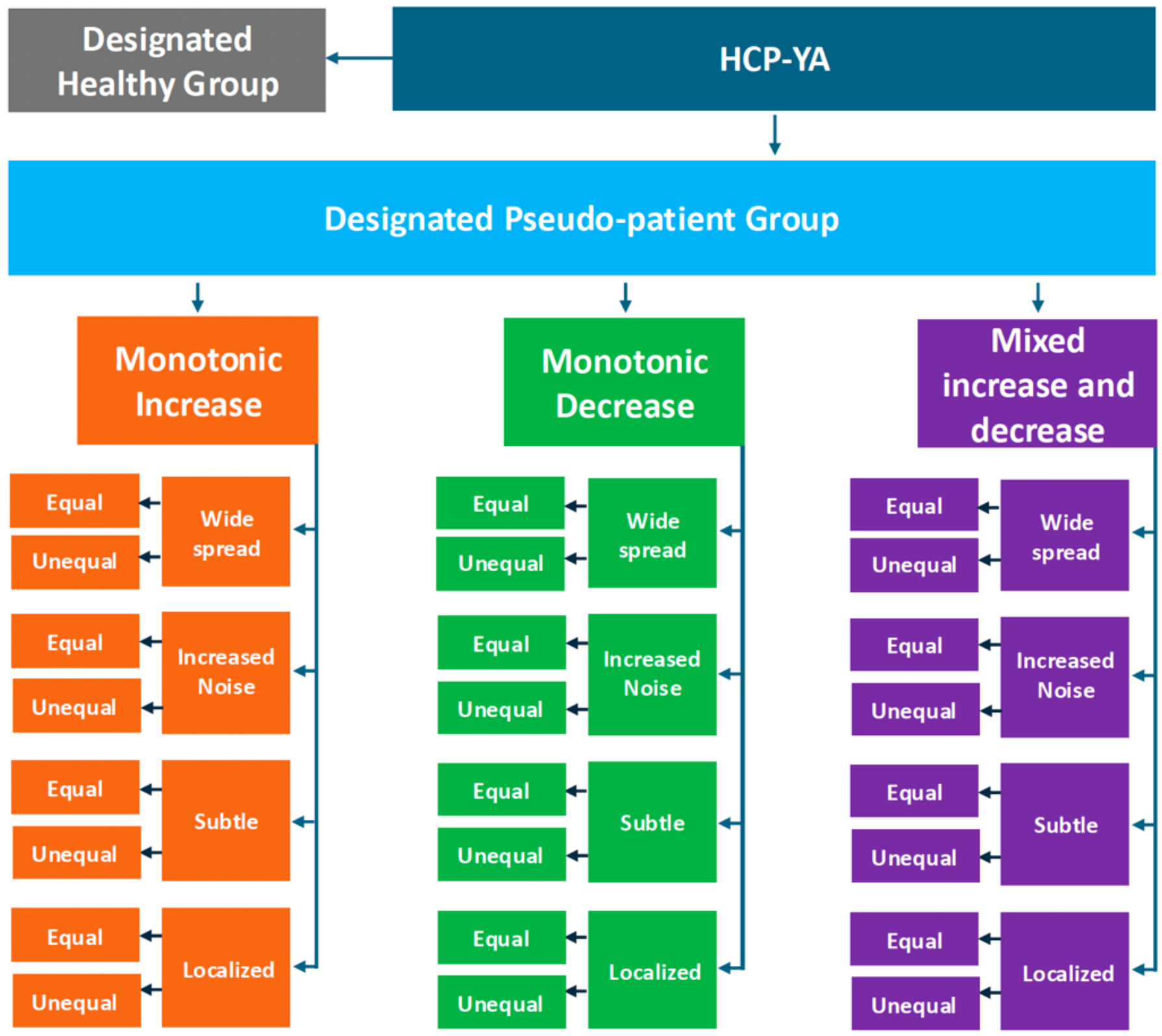

**Figure 1: Workflow of Synthetic datasets generation from the Human Connectome Project–Young Adult (HCP-YA) cohort**

The original neuroimaging data of the subgroup randomly designated as “healthy individuals” (n=508) remained unchanged throughout the study, providing a stable reference across conditions. The data for the group designated as pseudo-patients (n=600) underwent multiple synthetic transformations to simulate a wide range of realistic disease-related scenarios. Further details on the data transformations, parameter settings, and synthetic cohort construction can be found in Supplemental Information Table 1.

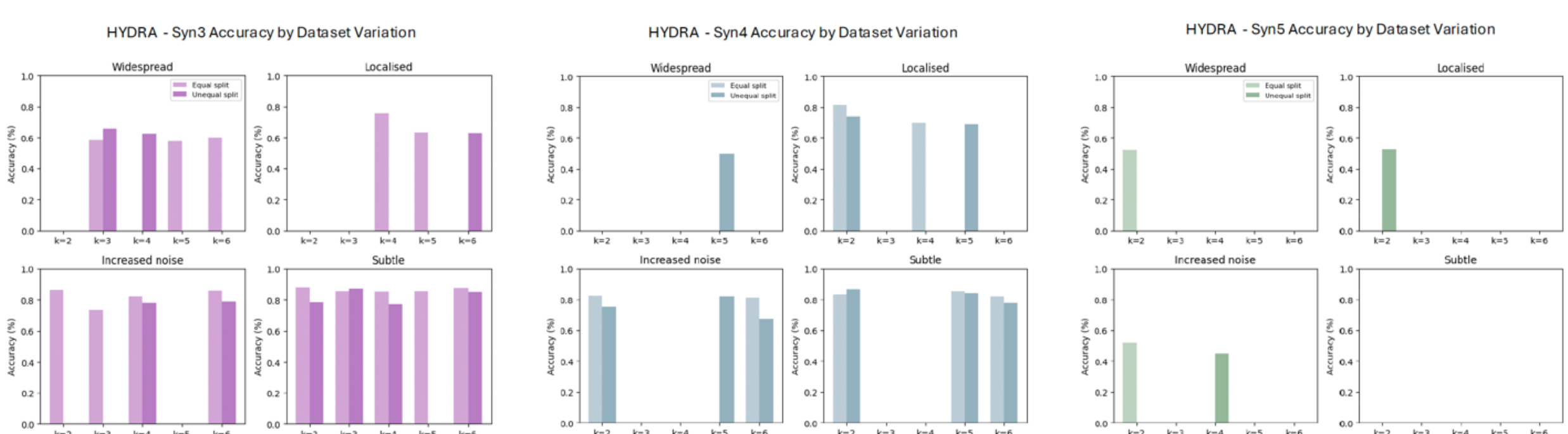

**Figure 2: HYDRA Accuracy Across Synthetic Datasets Syn3–Syn5**

Each subplot presents the accuracy of HYDRA on the Syn3, Syn4, and Syn5 synthetic datasets, evaluated across the four main data transformation types applied to the pseudo-patient group. Bars represent the accuracy relative to the known ground truth for varying numbers of clusters (k = 2 to 6), comparing performance under Equal and Unequal class distributions. HYDRA produces individual-level cluster assignments. Accuracy results are not displayed for conditions where the algorithm failed to generate valid clustering solutions, as no outputs were produced in those cases.

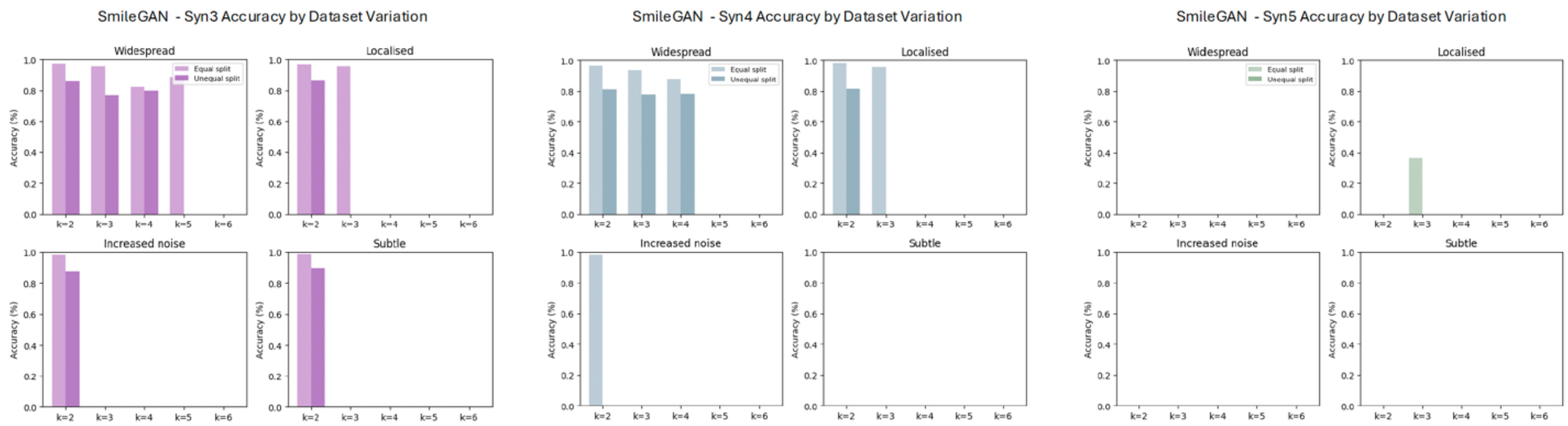

**Figure 3: SmileGAN Accuracy Across Synthetic Datasets Syn3–Syn5**
Each subplot presents the accuracy of SmileGAN on the Syn3, Syn4, and Syn5 synthetic datasets, evaluated across the four main data transformation types applied to the pseudo-patient group. Bars represent the accuracy relative to the known ground truth for varying numbers of clusters (k = 2 to 6), comparing performance under Equal and Unequal class distributions. SmileGAN clusters data into disorder-related patterns (R-indices). Accuracy results are not displayed for conditions where the algorithm failed to generate valid clustering solutions, as no outputs were produced in those cases.

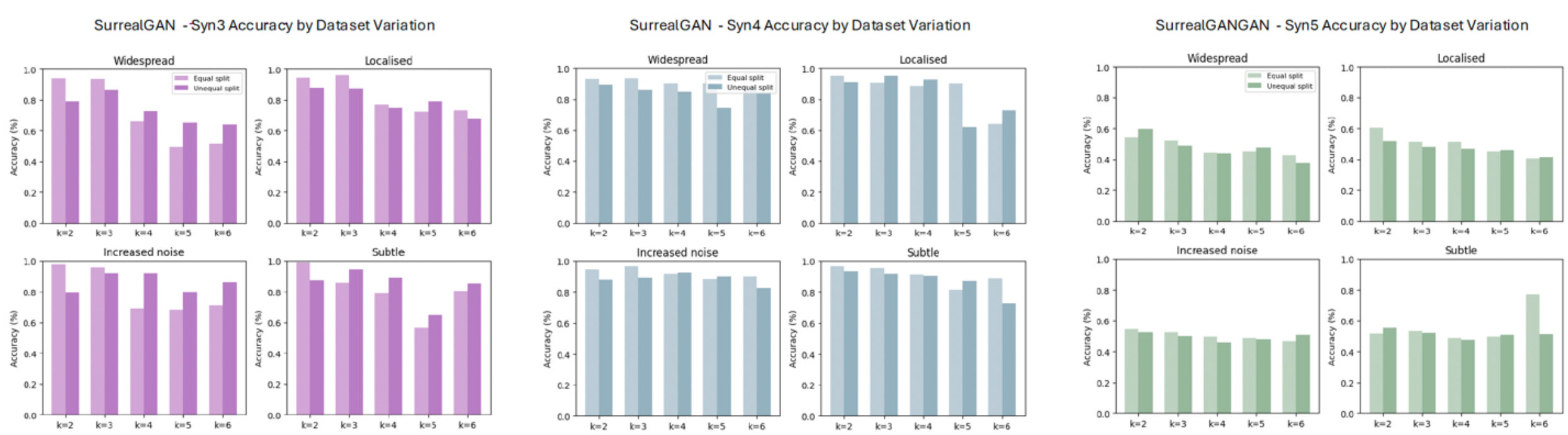

**Figure 4: SurrealGAN Accuracy Across Synthetic Datasets Syn3–Syn5**
Each subplot presents the accuracy of SurrealGAN on the Syn3, Syn4, and Syn5 synthetic datasets, evaluated across the four main data transformation types applied to the pseudo-patient group. Bars represent the accuracy relative to the known ground truth for varying numbers of clusters (k = 2 to 6), comparing performance under Equal and Unequal class distributions. SurrealGAN clusters data into disorder-related patterns.

Supplementary Information

*HeterogeneitY through DiscRiminant Analysis (HYDRA):* HYDRA is a non-linear, semi-supervised machine learning algorithm designed to integrate binary classification with subpopulation clustering (Valor et al., 2017) (Supplementary – Figure S1). It can accommodate two-dimensional vectors of any length and neuroimaging input features transformed into spatially meaningful "parcels: that serve as analytical units for modeling neuroimaging. this context, parcels refer either to distinct brain regions that can be derived using atlases (e.g., Desikan-Killiany, Schaefer, AAL) or data-driven brain structural patterns of shared variance derived from methods such as principal component analysis (PCA) or independent component analysis (ICA). The algorithm's codebase is available at https://github.com/anbai106/mlni. The core functionality of HYDRA involves separating a reference healthy control group from the clinical sample by constructing a convex polytope—a geometrical structure formed by combining multiple linear maximum-margin classifiers. These classifiers estimate hyperplanes that divide the data into distinct classes, ensuring that the margin (i.e., the distance between the decision boundary and the nearest data points from each class) is maximized. The identification of multiple dimensions of heterogeneity within the clinical population is achieved by varying the number of hyperplanes. To enhance the reliability and stability of the clustering results, HYDRA employs Determinantal Point Processes (DPPs) (Kulesza and Taskar, 2012). DPPs help ensure diverse and well-distributed initializations for clustering by selecting a subset of data points with high diversity in feature space. These diverse initializations are coupled with a multi-initialization strategy, generating multiple clustering solutions. The final clustering solution is derived through a consensus aggregation step, which synthesizes the outcomes of these multiple initializations into a unified solution.

*SeMI-supervised cLustEring via Generative Adversarial Network* (SmileGAN): Smile-GAN is a semi-supervised machine learning algorithm that leverages a generative adversarial network (GAN) architecture to cluster patients based on the patterns of their multivariate differences in reference to healthy controls (Yang et al., 2021) (Supplementary – Figure S1). The codebase for this algorithm is available at https://github.com/zhijian-yang/SmileGAN. The algorithm is applicable to any neuroimaging dataset, provided the data are transformed into "parcels" using brain atlases or statistical decomposition techniques. Before input into Smile-GAN, data from the patient group are z-transformed using the healthy individuals' group as a reference. The GAN architecture of Smile-GAN consists of typical generator and discriminator networks. The generator creates multiple mapping functions using control group data to generate pseudo-patient data that mimic the distribution of real patient data while maintaining consistency with the healthy control group structure. The discriminator employs an adversarial loss function to differentiate between real patient data and pseudo-patient data aiming to approximate the true underlying patterns in the patient data. The training process uses two hyperparameters to balance the relative importance of regularization terms, including change loss and cluster loss. Additionally, the algorithm incorporates the following forms of regularization during training to reduce noise: (a) Sparse transformations encourage prioritization of key features by the mapping functions, reducing noise and improving interpretability; (b) Lipschitz continuity to generate mapping functions that are smooth and stable and prevent extreme variations in predictions; and (c) Clustering function to identify distinct imaging patterns in the pseudo-patient data. This step is followed by an inverse mapping process to estimate the corresponding clusters in the real patient data.

*Semi-Supervised Representation Learning via GAN* (SurrealGAN): Surreal-GAN is a semi-supervised machine learning algorithm that decomposes disease-related brain changes into low-dimensional representations, referred to as R-indices (Supplementary – Figure S1). Each R-index captures a distinct deviation pattern between the patient group and the healthy control reference group (Yang et al., 2022). The codebase for this algorithm is available at https://github.com/zhijian-yang/SurrealGAN. The generation of R-indices relies on a transformation function that converts data from the healthy control group into pseudo-patient data, designed to replicate the distribution of real patient data. An inverse mapping function is then used to interpret the synthesized data. This function consists of decomposition, which extracts the R-indices, and reconstruction aimed at ensuring consistency between the original and inferred data. Surreal-GAN incorporates several regularization techniques and loss functions. Monotonicity loss is employed to ensure consistent and monotonic progression of disease-related deviations. Double sampling introduces variability during training to improve the model's generalizability, while orthogonality regularization aims at strengthening the separation between the different R-indices. Function continuity and transformation sparsity are applied to reduce noise and improve the interpretability of the results. Lastly, inverse consistency enhances the alignment and accuracy of the forward and inverse mappings aimed at improving the reliability of the R indices. While Surreal-GAN does not perform clustering directly, the R-indices serve as key features that define common characteristics across patients and can be used to identify homogeneous clusters of patients based on disease progression.

*Subtype and Stage Inference* (SuStaIn): SuStaIn is an unsupervised machine learning algorithm purporting to infer temporal patterns of brain changes from cross-sectional neuroimaging data (Young et al., 2018) (Supplementary – Figure S1). The codebase for SuStaIn is available at https://github.com/ucl-pond/pySuStaIn. This algorithm can be applied to any neuroimaging dataset, provided the data are transformed into "parcels" using brain atlases or statistical decomposition methods. SuStaIn employs event-based ordering (Young et al., 2015) to identify subgroups of patients who share similar patterns of disease progression. Before inputting data into the algorithm, patient group values for each parcel are z-transformed using the healthy control group as a reference. Disease progression is represented as a linear z-score model, where the progression is parameterized by a series of z-score "events." These events correspond to thresholds of change in specific parcels, and their sequence is assumed to follow a random monotonic order. To account for uncertainty in assigning patients to specific progression patterns, SuStaIn incorporates Markov Chain Monte Carlo (MCMC) sampling to generate probability estimates of patient assignment to progression patterns.

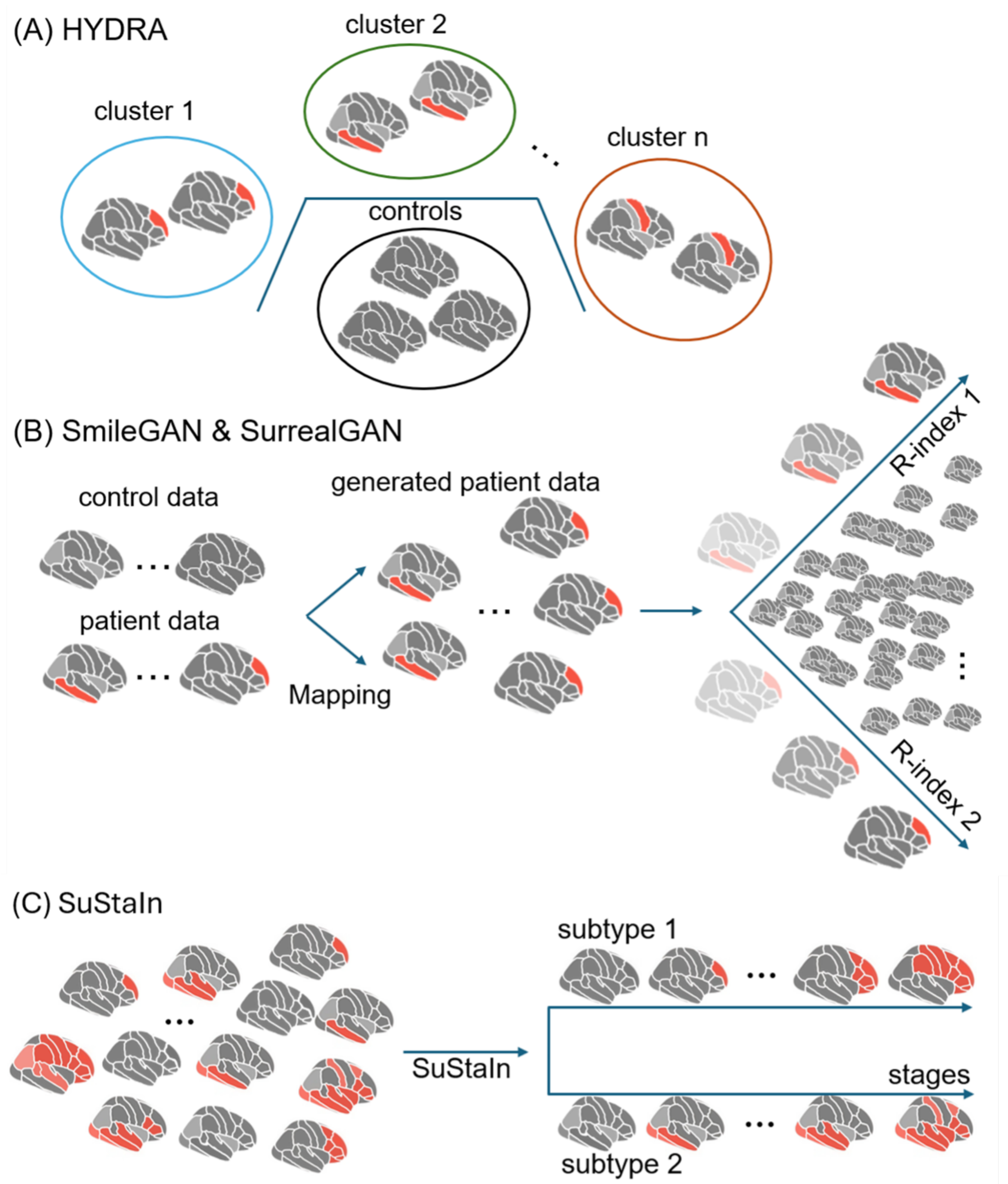

Supplementary - Figure S1

| **Supplementary Table S1. Three core datasets—Syn3, Syn4, and Syn5—were designed to simulate monotonic increases, monotonic decreases, or mixed random changes in morphometric values, respectively, with each serving as the basis for additional within-dataset variations detailed in the table.** | | | |
|---|---|---|---|
| **Within-Dataset Variation** | **Split Type** | **#Pseudo Patient Clusters** | **Size of Pseudo Patients' Clusters** |
| **Widespread**<br><br>21 variables per cluster, 6 overlap; σ = 0.05; α = 0.3 | Equal | 2 | 300 each |
| | Equal | 3 | 200 each |
| | Equal | 4 | 150 each |
| | Equal | 5 | 120 each |
| | Equal | 6 | 100 each |
| | Unequal | 2 | 200, 400 |
| | Unequal | 3 | 100, 200, 300 |
| | Unequal | 4 | 75, 125, 175, 225 |
| | Unequal | 5 | 20, 70, 120, 170, 220 |
| | Unequal | 6 | 25, 55, 85, 115, 145, 175 |
| **Increased Noise**<br><br>21 variables per cluster, 6 overlap; σ = 0.20; α = 0.3 | Equal | 2 | 300 each |
| | Equal | 3 | 200 each |
| | Equal | 4 | 150 each |
| | Equal | 5 | 120 each |
| | Equal | 6 | 100 each |
| | Unequal | 2 | 200, 400 |
| | Unequal | 3 | 100, 200, 300 |
| | Unequal | 4 | 75, 125, 175, 225 |
| | Unequal | 5 | 20, 70, 120, 170, 220 |
| | Unequal | 6 | 25, 55, 85, 115, 145, 175 |
| **Subtle**<br><br>21 variables per cluster, 6 overlap; σ = 0.05; α = 0.20 | Equal | 2 | 300 each |
| | Equal | 3 | 200 each |
| | Equal | 4 | 150 each |
| | Equal | 5 | 120 each |
| | Equal | 6 | 100 each |
| | Unequal | 2 | 200, 400 |
| | Unequal | 3 | 100, 200, 300 |
| | Unequal | 4 | 75, 125, 175, 225 |
| | Unequal | 5 | 20, 70, 120, 170, 220 |
| | Unequal | 6 | 25, 55, 85, 115, 145, 175 |
| **Localized**<br><br>6 variables per cluster; no overlap; σ = 0.05; α = 0.30 | Equal | 2 | 300 each |
| | Equal | 3 | 200 each |
| | Equal | 4 | 150 each |
| | Equal | 5 | 120 each |
| | Equal | 6 | 100 each |
| | Unequal | 2 | 200, 400 |
| | Unequal | 3 | 100, 200, 300 |

| **Supplementary Table S1. Three core datasets—Syn3, Syn4, and Syn5—were designed to simulate monotonic increases, monotonic decreases, or mixed random changes in morphometric values, respectively, with each serving as the basis for additional within-dataset variations detailed in the table.** | | | |
|---|---|---|---|
| **Within-Dataset Variation** | **Split Type** | **#Pseudo Patient Clusters** | **Size of Pseudo Patients' Clusters** |
| | Unequal | 4 | 75, 125, 175, 225 |
| | Unequal | 5 | 20, 70, 120, 170, 220 |
| | Unequal | 6 | 25, 55, 85, 115, 145, 175 |